\documentclass[conference,letterpaper]{IEEEtran}
\usepackage[letterpaper, margin=0.75in, top=1in]{geometry}

\IEEEoverridecommandlockouts
\usepackage{amsmath,amssymb,amsfonts}
\usepackage{algorithm2e}
\usepackage{graphicx}
\usepackage{hyperref}
\usepackage{textcomp}
\usepackage{xcolor}
\usepackage{xspace}
\usepackage{booktabs}
\usepackage{multirow}
\usepackage[explicit]{titlesec}
\usepackage{subcaption}
\titleformat{\paragraph}[runin]{\normalfont\normalsize\bfseries\scshape}{}{0pt}{#1}
\titlespacing*{\paragraph}{0pt}{1ex}{1ex}

\usepackage[backend=biber, style=ieee]{biblatex} 
\newcommand{\Loss}{\mathcal{L}}

\DeclareMathOperator{\E}{\mathbb{E}}

\newcommand{\modellong}{Behavioral Cloning Value Approximation\xspace}
\newcommand{\modelshort}{\textit{BCVA}\xspace}

\begin{document}

\title{
Asking for Help: Failure Prediction in Behavioral Cloning through Value Approximation
}

\author{\IEEEauthorblockN{Cem Gokmen\textsuperscript{*}}
\IEEEauthorblockA{\textit{Stanford University} \\
Stanford, CA \\
cgokmen@stanford.edu}
\and
\IEEEauthorblockN{Daniel Ho\textsuperscript{\dag}}
\IEEEauthorblockA{\href{https://www.everydayrobots.com/}{\textit{Everyday Robots}} \\
Mountain View, CA \\
danielho@waymo.com}
\and
\IEEEauthorblockN{Mohi Khansari}
\IEEEauthorblockA{\href{https://www.everydayrobots.com/}{\textit{Everyday Robots}} \\
Mountain View, CA \\
khansari@everydayrobots.com
\thanks{\textsuperscript{*}Work done while at \href{https://www.everydayrobots.com/}{Everyday Robots}.}
\thanks{\textsuperscript{\dag}Now at Waymo.}
}
}

\maketitle

\begin{abstract}
Recent progress in end-to-end Imitation Learning approaches has shown promising results and generalization capabilities on mobile manipulation tasks. Such models are seeing increasing deployment in real-world settings, where scaling up requires robots to be able to operate with high autonomy, i.e. requiring as little human supervision as possible. In order to avoid the need for one-on-one human supervision, robots need to be able to detect and prevent policy failures ahead of time, and \textit{ask for help}, allowing a remote operator to supervise multiple robots and help when needed. However, the black-box nature of end-to-end Imitation Learning models such as Behavioral Cloning, as well as the lack of an explicit state-value representation, make it difficult to predict failures. To this end, we introduce \modellong (\modelshort), an approach to learning a state value function based on and trained jointly with a Behavioral Cloning policy that can be used to predict failures. We demonstrate the effectiveness of \modelshort by applying it to the challenging mobile manipulation task of latched-door opening, showing that we can identify failure scenarios with with $86\%$ precision and $81\%$ recall, evaluated on over 2000 real world runs, improving upon the baseline of simple failure classification by $10$ percentage-points.
\end{abstract}


\section{Introduction}
The field of robotics has seen significant developments in recent years on mobile manipulation tasks. Improved methods have made it possible to learn from simulated and real-world experiences together~\cite{tcl}. Being able to fuse data from different sensor modalities, such as RGB and depth, has resulted in greatly improved action-taking decisions and thus manipulation behaviors~\cite{vib}. With these improvements, it has become possible to deploy such mobile manipulation agents in the real world to solve practical real-life tasks. In this context, Imitation Learning (IL) approaches such as Behavioral Cloning have been established as a practical solution to many mobile manipulation tasks, such as \textit{latched door opening}.

\begin{figure}[t!]
    \centering
    \begin{subfigure}{\linewidth}
        \includegraphics[width=\linewidth]{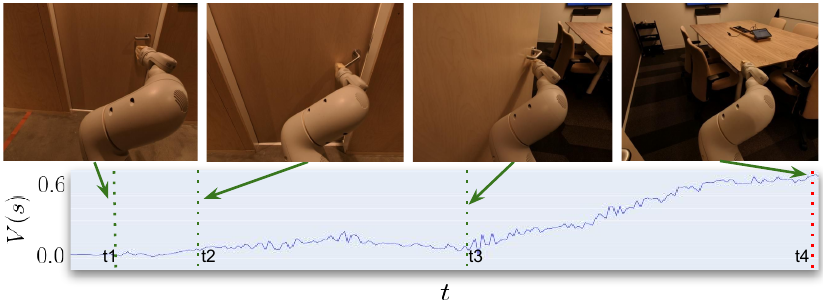}
        \vspace{-5mm}
        \caption{An example of a successful autonomous latched door opening run. At the beginning the estimated state value is zero; but, as the robot progresses towards the end of the task, the state value increases, indicating that the robot is going to finish the task successfully.}
        \vspace{2mm}
    \end{subfigure}
    \begin{subfigure}{\linewidth}
        \includegraphics[width=\linewidth]{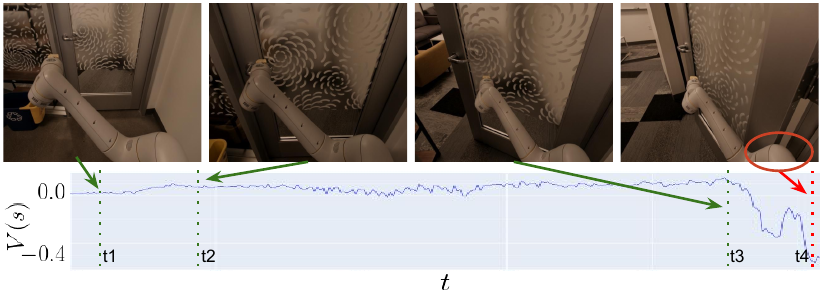}
        \vspace{-5mm}
        \caption{An example of a failed policy run. Similar to above, the value at the beginning of the episode is near zero. As the robot swings the door open, the value remains positive. However, at time $t_4$, the robot elbow is about to collide with the door frame, and the state value drops to below -0.3, upon which the policy stops and asks for help.}
    \end{subfigure}
    \caption{Running an autonomous latched door opening policy on an Everyday Robots mobile manipulator. By using BCVA, the policy can predict potential failures in the near future, stop, and ask for help.}
    \vspace{-3mm}
    \label{fig:ask_for_help}
\end{figure}

The ideal procedure for deploying Imitation Learning models in the real world would consist of two phases: a \textit{training phase} where a human operator performs demonstrations in order to learn a policy, and an \textit{operational phase} where the robot can operate, ideally unattended, in an open environment to perform its task. However, Imitation Learning approaches have some shortcomings that make their naive implementation problematic for such real-world deployments. As an important example, agents trained using this paradigm tend to perform poorly in out-of-distribution states. Therefore, in the operational phase, the policy will continue to execute with compounding error under out-of-distribution states until failure can be externally detected post-factum, e.g. by a human operator, a sensor, etc., which can lead to damage to the robot and the environment. The need to prevent these catastrophic failures means the robots need continuous one-on-one human supervision, blurring the lines between the training and operational phases. 

One straightforward solution to this problem is to make sure as much of the robot's state space is covered by the training data as possible. In the training phase where one-on-one supervision is available, this can be achieved by applying data collection regimens, such as DAgger~\cite{dagger}, that iteratively include more of the state space in the training distribution. However, even after applying DAgger, it is inevitable that the robot will encounter scenarios during deployment that it has not seen in training. In this case, it is desirable for the robot policy itself to be able to identify that it is going to be unable to solve the task successfully. It can then preemptively stop and ask for help, avoiding any damage to the environment and allowing human experts to provide corrective demonstrations. This mode of operation also allows the robots to operate outside of one-on-one human-to-robot supervision, by allowing \textbf{remote} supervision where human operators at an off-site control center can loosely monitor multiple robots and intervene only when asked for help. This allows retraining in the real world in an \textit{incremental} manner: initial policies from the training phase are deployed on real robots, additional data is collected during the operational phase through both regular operations and expert corrections, and the policies are re-learned periodically from the continuously-growing dataset.

However, in order to predict failure, the robot needs to quantify its confidence in its ability to solve the task given its current state. While this role is traditionally filled by a state-value function in the Reinforcement Learning paradigm, Behavioral Cloning does not provide an explicit value function representation. One way of filling this gap is by applying Reinforcement Learning approaches alongside Behavioral Cloning as in~\cite{saycan2022arxiv} purely for the purpose of obtaining a state-value representation, which comes at great computational cost and low data efficiency.

To this end, we introduce our main contribution, \modellong (\modelshort): a method for computing state-value representations conditioned by a history of policies learned through incremental updates to the training data and the resulting evaluations from said policies. \modelshort has three main advantages:

\begin{itemize}
    \item It allows the state values to be batch-computed offline using policy evaluation data (and variable reward discounting regimes) prior to training, rather than through exploration in expensive paradigms such as Reinforcement Learning.
    \item It can be learned simply as a regression head on top of an existing Behavioral Cloning model, allowing for low-cost training and inference as well as weight sharing.
    \item When trained jointly with the Behavioral Cloning model, it allows failure examples to also be used for representation learning, increasing the data available for learning a state embedding that helps the policy generalize better.
\end{itemize}

To test our approach, we deploy \modelshort on a mobile manipulation robot solving the \textit{latched door opening} task, where the robot must approach a closed latched door, grasp the handle and rotate it to release the latch, drive forward to open the door, and enter the room. This task is challenging due to the large number of failure modes that stem from minor errors, e.g. being unable to find or grasp the handle, grasping it too close to the pivot point to be able to apply the necessary force, or colliding with the door frame or the door itself. We show that \modelshort deployed together with an 85\% successful end-to-end BC policy based on \cite{vib} is able to correctly identify and request help on 81\% of the policy failures across all of the aforementioned failure modes, while incorrectly requesting help (false-positive) on as little as 14\% of successful trajectories.

\section{Related Work}
\subsection*{Failure Detection / Prediction}
\paragraph{On Structured Data} The problem of Failure Detection and Prediction in robotics has received considerable interest in the past decades. \cite{Diryag2014NeuralNF} introduced the first Neural Network-based robot fault prediction algorithm using a model trained on force and torque sensor data, which other works such as \cite{AlonsoTovar2016BayesianNC}, \cite{Li2014AnIK} and \cite{Liu2019DeepCN} expanded on. The common theme of these methods is that they are 1) policy-agnostic, in that they do not share weights or data with the robot policy, and 2) sensor-based, in that they use structured numeric data from a range of robot sensors to complete prediction, therefore avoiding the problem of representation learning on unstructured sensor data, e.g. images, as in our case. As a result, they are able to simply classify success/prediction from time series data as in our classification baseline with high success, an approach that does not transfer well to the more complex problem of in-the-wild failure detection from vision data.

\paragraph{In Reinforcement Learning} In the context of Reinforcement Learning, asking for help and failure prediction can be achieved by thresholding based on policy-conditioned state values. To compute state values for a given policy (a task known as Policy Evaluation), a large variety of methods have been proposed. Naive Policy Evaluation~\cite{sutton2018reinforcement}, which relies on generating rollouts of the given policy and using dynamic programming methods to propagate terminal rewards to intermediate states, can be used when it is possible to generate rollouts using the target policy. In the context of mobile manipulation, however, this can only be done in simulation, where most policies tend to perform much better than they do in reality due to the simulation-to-reality gap. Another branch of the literature, known as Off-Policy Policy Evaluation, focuses on the very relevant scenario of evaluating the performance of a new (\textit{target}) policy using data collected using other (\textit{behavior}) policies. This problem has been studied in depth, and a number of methods such as \cite{Precup2000EligibilityTF}, \cite{Precup2001OffPolicyTD}, \cite{Thomas2016DataEfficientOP} and \cite{Kallus2020DoubleRL} as well as benchmarks such as \cite{Fu2021BenchmarksFD} and \cite{Voloshin2021EmpiricalSO} have been introduced. However, these methods are difficult to apply in the context of Partially-Observable Markov Decision Processes (POMDPs) such as vision-based mobile manipulation since they largely rely on tabular data and the presence of explicit transition models, even if approximate, which makes them incompatible with Behavioral Cloning.

\paragraph{In Imitation Learning} Similarly, the problem of detecting success/failure probabilities has been studied in the context of Imitation Learning policies. One line of work focuses on detecting out-of-distribution states to prevent robot failures. \cite{Richter2017SafeVN}, \cite{Hirose2018GONetAS} and \cite{momart} rely on generative methods and reconstruction errors as a measure of uncertainty, whereas \cite{ensembledagger} uses the divergence of an ensemble of policies for uncertainty estimation. While these methods perform well in the task of detecting out-of-distribution scenarios, the detection of out-of-distribution scenarios itself does not necessarily constitute the most appropriate proxy task for error detection: many times, forward prediction can still be successfully performed under known failure conditions, and other times, high-entropy states (e.g. where objects are occluded or colliding etc.) where forward prediction is inherently difficult can show high reconstruction errors despite there being no failure. These methods also do not make use of existing evaluation data from previous behavior policies, a valuable stream of data in the context of iterative policy training. Other works such as \cite{saycan2022arxiv} apply temporal difference-based Reinforcement Learning methods alongside Behavioral Cloning to learn explicit value functions for imitation-based policies. While effective, this approach is prohibitively costly in terms of computational and data efficiency, since it relies on generating and evaluating on-policy rollouts for the target policy.

\subsection*{Dataset Collection}
A related problem in the context of Imitation Learning is Dataset Collection, i.e. how to collect the right demonstrations that allow a Behavioral Cloning policy to improve on failure scenarios with the highest possible data efficiency. To this end, DAgger~\cite{dagger} is a well-tested mechanism that can be used at the time of data collection (in the \textit{training phase}) to make sure that frequently-encountered but out-of-training-distribution states get new labelled data to help the policy recover from such states. However, the utility of DAgger in the case of mobile manipulation is limited, not only due to the high dimensionality of the state space and the low efficiency of collecting per-frame corrections from the experts, but also due to the simple fact that in a real-world deployment, many error scenarios will be encountered for the first time in the \textit{operational} phase where the robots need to operate mostly unsupervised. In this context, even though corrective expert demonstrations can be collected post-failure, the DAgger strategy still relies on failing first and then learning to recover, rather than learning to stop prior to the failure. Furthermore, even when robots can be supervised continuously, when humans are instructed to preemptively stop the robot before failure, empirical evidence shows that they are inclined to intervene either too early (e.g. stopping a successful rollout because the robot's non-human embodiment makes the rollout feel unfamiliar) or too late (e.g. the robot quickly goes from an acceptable to a high-risk state without chance for intervention). Altogether, correcting naive imitation learning approaches with a DAgger-like strategy during deployment requires costly around-the-clock human operator monitoring, which nonetheless does not provide a simple and reliable mechanism for providing demonstrations to learn to avoid known failures.

\subsection*{Deep Learning for Mobile Manipulation}
Since our approach relies on jointly training a policy and a value approximation, it needs to be built upon a reliable framework for autonomous mobile manipulation. While autonomous mobile manipulation has been studied as early as the 1990s~\cite{oldMM} even for our chosen problem of door opening~\cite{dooropeningIL}, early solutions were limited in their ability to generalize to diverse environments and long-horizon tasks. More recently, Deep Learning-based methods have performed imitation learning on RGB and Depth sensor observations, such as in \cite{newMMIL} and \cite{newMMIL2}. Furthermore, works such as \cite{BCzero} have demonstrated that vision-based Behavioral Cloning can be applied to long-horizon tasks. For this work, we build particularly upon our previous works in \cite{tcl}, where we apply a novel loss to both simulation and real-world data to train a Behavioral Cloning policy, and \cite{vib}, where we use Variational Inference methods to encourage generalization in Behavioral Cloning. We discuss the particulars of our policy model in the Methods section.

\section{Problem Setup}
Given a mobile manipulation task, our goal is to \textit{jointly} learn:
\begin{itemize}
    \item a policy $\pi(a|s)$ that outputs an action $a$ (e.g. a joint velocity) given a state $s$ (e.g. RGB and depth images) in order to complete the task,
    \item a policy-conditioned state value function $V_\pi(s)$ that assigns a higher value to states that have a higher probability of resulting in a successful rollout.
\end{itemize}

For learning the policy, we assume that we have a dataset $D^* = \{\tau^*_0, \tau^*_1, ..., \tau^*_N\}$ consisting entirely of expert demonstrations $\tau^* = (s_0, a_0, s_1, a_1, ..., s_{T-1}, a_{T-1}, s_T)$ that result in successful completion of the task at state $s_T$, with actions generated by an expert policy $\pi^*$. We use Behavioral Cloning to learn to imitate this policy, where the objective is to minimize the divergence between our policy $\pi(a|s)$ and the expert policy $\pi^*(a|s)$.

For learning the value function, it is necessary to also have trajectories of scenarios where the policy fails to complete the task. Our approach focuses specifically on \textit{incremental} robot learning scenarios where policies are periodically evaluated and re-learned using more training data. We assume that we start with an initial policy $\pi_0(a|s)$ trained on an initial dataset of demonstrations $D_0^*$ with no value estimate. A dataset of rollouts from the first policy then needs to be collected and labelled under full human supervision, and discounted returns computed offline to be used in bootstrapping the value estimate. Then, each following policy $\pi_k(a|s)$ is trained jointly with a value estimate $V_k(s)$ from previous episodes' labelled rollouts. The policy is then deployed on a robot in the real world, where the robot repeatedly performs the task during daily operations, generating trajectories $\tau^{\pi_k}_i$. All the trajectories during operation are being recorded irrespective of whether or not the trajectory ends with a success ($r=1$) or failure ($r=-1$). Human operators are also allowed to provide additional expert demonstrations $D^*_k$ for failure cases and help requests by the robot. These demonstrations are then added to the existing dataset to create the training set for the next policy $\pi_{k+1}$: $D_{k+1} = D_k \cup {D^*_k} \cup \{\tau^{\pi_k}_1, \tau^{\pi_k}_2, ..., \tau^{\pi_k}_i\}$. This dataset aggregation mechanism is described in detail in Algorithm~\ref{alg:dataset}.

\LinesNumbered
\begin{algorithm}[t!]
  \caption{Using \modelshort to continuously improve the model performance through dataset aggregation of both expert demonstrations and autonomous policy rollouts.}
  \label{alg:dataset}
  $D \gets$ initial expert demonstrations\;
  $\pi \gets train(D)$ \;
    \While{labelling for value bootstrapping continues}
    {
            $\tau \gets rollout(\pi)$ \;
            $D \gets D \cup \{\tau$\} \;
    }
    \While{operational deployment continues}
    {
        $\pi, V \gets train(D)$ \;
        \While{operational cycle for $\pi$ continues}{
            $\tau \gets rollout(\pi, V)$ \;
            $D \gets D \cup \{\tau$\} \;
            \If{$\tau$ includes an ask for help}{
                $\tau^*_{new} \gets$ new expert demonstration\;
                $D \gets D \cup \{\tau^*_{new}$\} \;
            }
        }
    }
    \vspace{2mm}
\end{algorithm}

\section{Method}
\label{sec:method}

\begin{figure*}[thpb]
    \centering
    \includegraphics[width=0.7\linewidth]{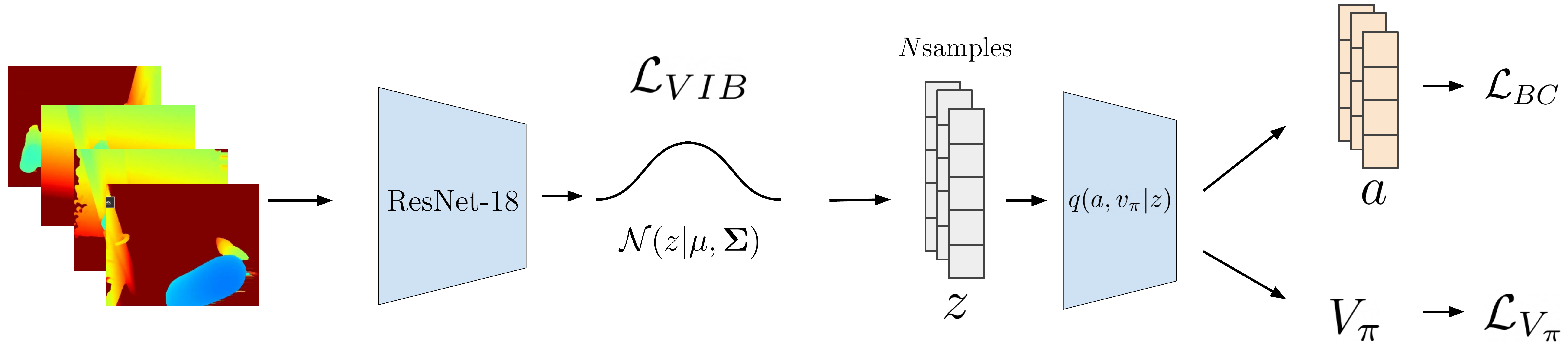}
    \caption{
    Model architecture overview. We jointly learn a policy $\pi(a|s)$ and an approximate state value $V(s)$. Similarly to \cite{vib}, Our model is agnostic to the choice of sensory input. However, in this paper we report on the results from a depth-only model since it results in the higher BC performance as described in \cite{vib}.}
    \label{fig:model}
\end{figure*}

\subsection{Policy Evaluation}
Given each state $s_j$ from trajectory $\tau_i$ in the dataset $D_k$, the matching terminal reward $r_i$, and a discount factor $\gamma$ accounting for how far in the future the reward is, we compute the discounted return function $G(s)$ similar to that in Monte Carlo methods:

\begin{equation}
G_\tau(s_j) = \gamma^{\Delta_j} r_i
\end{equation}

\noindent where $\Delta_j$ is the accumulated ``distance'' from $s_j$ to the final state of the trajectory $s_{|\tau_i|}$ using a distance function $\delta(s_a, s_b)$:

\begin{equation}
\Delta_j = \sum_{j \leq t < |\tau_i|} \delta(s_t, s_{t+1})
\end{equation}

A simple distance function is the time difference between two states ($\delta(s_a, s_b) = b - a$), in which case the discounted reward is equivalent to the discounted reward in a Markov Decision Process:

\vspace{-2mm}
\begin{equation}
G_\tau(s_j) = \gamma^{|\tau_i| - j} r_i
\end{equation}

In this paper, we also consider two other distance functions: 1) the normalized difference between the pixels in the input space, and 2) the normalized difference in the robot's kinematic state.
For the pixel difference, we define our distance function as the normalized sum of the absolute pixel differences:

\vspace{-2mm}
\begin{equation}
\delta(s_a, s_b) = \frac{(\sum_x^w \sum_y^h \sum_c^C | s_b(x, y, c) - s_a(x, y, c) |) - \tilde \mu}{\tilde \sigma} + 0.5
\end{equation}

\noindent where $w$ is the RGB sensor width, $h$ the RGB sensor height, $C$ the number of channels, and $s_t(x, y, c)$ the value of the $c$'th channel of the pixel at $(x, y)$ of state $s_t$. The sample mean $\tilde \mu$ and standard deviation $\tilde \sigma$ are used to scale the values such that one standard deviation falls in the $(0.5, 1.5)$ range, suitable for use as the discount factor exponent.

Similarly, we compute the kinematic state difference as the normalized sum of the joint states and robot base pose:

\begin{align}
\begin{split}
\delta(s_a, s_b) &= \frac{(\sum_{j \in J} |s_b(j) - s_a(j)|) - \tilde \mu_{joint}}{2 \tilde \sigma_{joint}} \\
&+ \frac{(\sum_{c \in xyz} |s_b(c) - s_a(c)|) - \tilde \mu_{xyz}}{2 \tilde \sigma_{xyz}} + 0.5
\end{split}
\end{align}

\noindent where $J$ is the set of all joints of the robot, $s(j)$ indicates the angular position of joint $j$, and $s(c)$ indicates the Cartesian velocity of the robot base along world axis $c$. Here, the joint angles and Cartesian coordinates are scaled separately since they have different units.

We compare the results of these different approaches in our experiments section.

\subsection{Policy / Value Approximation Network}
Our goal is to jointly learn a policy $\pi(a|s)$ and an approximate state-value $V(s)$, from our dataset consisting of expert demonstrations and policy evaluations, where the policy replicates the expert actions and the state value approximates the average discounted returns. Following the Bayesian Imitation Learning method described in \cite{vib}, we combine a stochastic state encoder network $p(z|s)$ with two decoder networks: an action decoder $q_a(a|z)$ and a value decoder $q_v(v|z)$. 

\paragraph{Policy Learning} To learn a policy that will replicate the expert actions, we apply a Huber loss~\cite{Huber} between the demonstrated actions $a$ and predictions $\hat{a}$ obtained through Monte Carlo sampling from our neural network decoder, following~\cite{BCzero}, as our Behavioral Cloning loss:

\begin{equation}
\Loss_{BC} = \E_{z \sim p(z|s)}[\E_{\hat{a} \sim q_a(\hat{a}|z)}[\Loss_{Huber}(a - \hat{a})]]
\end{equation}

\noindent where $a \in \mathcal{R}^{10}$ is the action command that includes: 2 DoF for the robot base, 9 DoF for the robot arm, and 1 DoF for the task termination prediction.

The Behavioral Cloning loss is only applied to training instances that are expert demonstrations, e.g. evaluation instances from previous policies are not used for policy learning.

\paragraph{Value Approximation} With the discounted returns $G_\tau(s)$ available from policy evaluation, state values in the discrete-state context can be computed by averaging the discounted returns over each time the state is visited, where $T_s$ is the set of trajectories that visit state $s$:

\begin{equation}
V(s) = \frac{1}{|T_s|} \sum_{\tau \in T_s} G_\tau(s)
\end{equation}

However, since our state representation (RGB or depth images) is continuous, each state $s$ appears typically in exactly one trajectory; but, we want our learned state value information to be shared between similar states. To this end, we use the stochastic value decoder $q_v(v | s)$ in our network, and we once again apply a Huber loss between Monte Carlo samples of predicted values and the discounted returns computed through policy evaluation:

\begin{equation}
\Loss_{V_\pi} = \E_{z \sim p(z|s)}[\E_{\hat{v} \sim q_v(\hat{v}|z)}[\Loss_{Huber}(v - \hat{v})]]
\end{equation}

Training instances both from human demonstrations and on-policy operations are used when computing this loss. The former includes only successful runs while the latter contains both successful and failed episodes.

\paragraph{Bottleneck} To discourage our model from overfitting, following \cite{vib}, we apply the Variational Information Bottleneck~\cite{originalvib} on the learned stochastic encoding $z$. $\Loss_{BC}$ and $\Loss_{V_\pi}$ together encourage $z$ to be maximally predictive of $a$ and $v$, respectively, and to encourage it to be minimally informative of $s$, we apply a Kullback–Leibler divergence loss $\Loss_{KL}$ between the state embedding posterior $p(z|s)$ and the learned prior $r(z)$. Due to the computational intractability of computing the KL divergence directly, we use Monte Carlo sampling to draw samples from both distributions and compute the sample divergences instead.

\paragraph{Combined Loss} To achieve all three objectives at once, we combine our losses into a single loss, weighted by two parameters, $\lambda$ to control the policy / value learning tradeoff, and $\beta$ to control the bottlenecking tradeoff:

\begin{equation}
\Loss = \Loss_{BC} + \lambda\Loss_{VA} + \beta\Loss_{KL}
\end{equation}

In our experiments, we heuristically chose $\lambda = 0.5$ and $\beta = 10^{-6}$.

\paragraph{Network Details} We use a multivariate Gaussian distribution on $\mathcal{R}^{256}$ as our stochastic encoder $p(z|s)$, parameterized by a ResNet-18~\cite{resnet} convolutional network that outputs the mean and covariance of the distribution. The action decoder $q_a(a|z)$ is a 2-layer MLP and the value decoder $q_v(v|z)$ a 3-layer MLP. The learned prior $r(z)$ is a multivariate Gaussian mixture with 512 components and learnable parameters.

\subsection{Asking for Help}
With our model able to output the policy action $a_t$ and the state value $V(s_t)$ given a state $s_t$, we follow a simple criteria to decide when to ask for help:

\begin{enumerate}
    \item If the state value $V(s_t) < \epsilon$, for some constant $\epsilon$, for more than some constant $\nu$ past frames, we stop and ask for help. (See experiments section for how to pick $\epsilon, \nu$)
    \item Otherwise, we continue by executing $a_t$, observing $s_{t+1}$, and re-evaluating the asking-for-help criteria.
\end{enumerate}

For operational deployment, we tune the thresholds $\epsilon$ and $\nu$ by running the value estimate on rollouts from the (human-labelled) validation set, computing the episode-level confusion matrix across different values of $\epsilon$ and $\nu$, and picking appropriate values such that the model satisfies requirements in terms of both overall precision and recall, as well as being able to correctly flag a small, hand-selected sample of failures of concern as described in Section \ref{sec:exp}.

\section{Experiments}
\label{sec:exp}
We train our model on a real-world dataset of $\sim2900$ expert demonstrations and bootstrap our value estimate with $\sim9000$ episodes of policy rollouts under a fully-supervised human operator setting. All expert demonstrations were success cases, whereas each of the policy rollouts was manually labelled as success or failure at the end of the episode, and discounted rewards for each step calculated offline post-factum. The policy rollouts were executed and labelled in a span of multiple months using 100+ different Behavioral Cloning models each independently trained with the action decoder head only. In this work, we use $100\%$ of the expert demonstrations and $75\%$ of the policy rollouts to train the proposed BCVA model. We use the remaining $25\%$ as a held-out validation dataset to evaluate model performance.

Both expert demonstrations and policy rollouts were taken using a fleet of 5+ Everyday Robots and operated across 50+ meeting rooms located in 4 different Alphabet buildings across the Mountain View campus. Please note that we did not modify the environment and the robots operate on rooms in the natural state they were left (see Figure \ref{fig:ask_for_help}).

We evaluate the performance of four different models on the asking for help scenario, including three variants of the proposed approach and a baseline using the classification head, similarly to \cite{BCzero}. As described in Section \ref{sec:method}, we propose three different distance functions to compute the discounted return: 1) time, 2) movement, and 3) pixel. We call each of these approaches BCVA-Time, BCVA-Movement, and BCVA-Pixel respectively.

Table \ref{tab:eval_comp} details the evaluation performance of each of the four models. All three BCVA variants outperform the classifier approach by at least $8$ percentage points. By using the value function, the model can better assign reward to each frame in the episode, compared to the classifier approach in which all frames get equal reward/label. 

Among BCVA variants, using time as a distance metric leads to marginally lower performance. This might be due to difficulty dealing with idling frames, where the robot is stuck or standing still, but the model still needs to predict different values for consecutive, look-alike image frames. The pixel distance function performs slightly better than movement distance function, probably because it can better account for the dynamic nature of the task, e.g. the door bouncing back after hitting the wall.

\begin{table}
  \centering
  \footnotesize{
  \begin{tabular}{@{}l|c|c|cc@{}}
    \toprule
    Method                & F1 & Accuracy & Precision & Recall\\
    \midrule
    Classifier (Baseline)  & 0.73 & 0.82 & 0.73 & 0.74 \\
    BCVA - Time (ours)     & 0.81 & 0.87 & 0.82 & 0.81 \\
    BCVA - Movement (ours) & 0.82 & 0.88 & 0.84 & 0.79 \\
    BCVA - Pixel (ours)    & \textbf{0.83} & 0.89 & 0.86 & 0.81 \\
    \midrule
    \bottomrule
  \end{tabular}
  }
  \caption{
  Performance comparison of four different asking-for-help methods. The values are calculated based on analysis of 2000+ real world policy rollouts.
  }
  \label{tab:eval_comp}
  \vspace{-0.5cm}
\end{table}

\begin{figure}[thpb]
    \begin{minipage}{\linewidth}
    \centering
    \includegraphics[width=\linewidth]{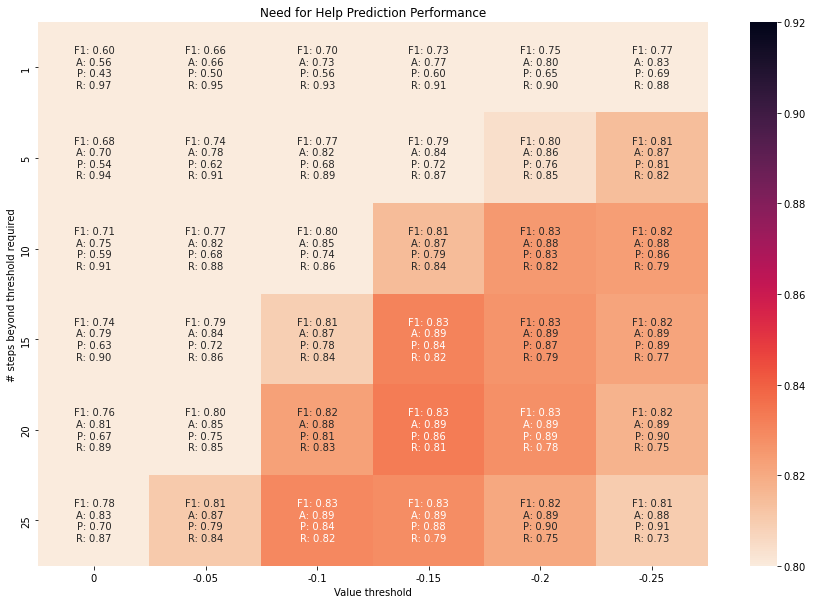}
    \caption{Illustration of the performance confusion matrix for different values of $\epsilon$ ($x$-axis) and $\nu$ ($y$-axis) calculated for the BCVA-Pixel model. The best model performance, an F1 score of 0.83, occurs when $\nu=20$ and $\epsilon = -0.15$.}
    \label{fig:eval_pixel_autonomous}
    \end{minipage}
\end{figure}

Figure \ref{fig:eval_pixel_autonomous} shows an example of a confusion matrix generated for the BCVA-Pixel method for different values of $\nu$ and $\epsilon$. Based on this confusion matrix, the best model performance occurs when $\nu=20$ and $\epsilon = -0.15$, which indicates the robot should ask for help after seeing 20 consecutive states (i.e. 2.0 sec) with the state-value less than or equal to $-0.15$. If it is desired to do prediction faster, e.g. in 0.5 second, there will be slight degradation in the prediction score; going from an F1 score of 0.83 to 0.81. Similarly, there is also a trade-off between precision-recall metric. Choosing a recall score of 0.97 (i.e. almost detecting most of the failure cases) comes at the cost of having $57\%$ false positives. 

\section{Conclusion}
We presented \modelshort, a method of joint Behavioral Cloning and state value estimation, which allows for scaling up the real-world deployment of Imitation Learning. Enabling  robots to ask for help removes the need to have continuous one-on-one human-robot supervision, which would thus reduce the cost and barriers of operations. In this new paradigm, $N$ robots, where $N > 1$, can run autonomously while being supervised by one (remote) operator, who can now focus solely on the robots that are asking for help. We tested \modelshort on a complex mobile-manipulation task of latched door opening with rich environment observations and contacts, and our model was able to identify failure risks with $86\%$ precision and $81\%$ recall. This ability is relatively straightforward to implement for typical neural-network-based Behavioral Cloning policies, as a separate prediction head on top of a shared feature-encoder backbone. Weight sharing allows for both policy learning and state value estimation to jointly improve and cue on shared features (e.g. leveraging failed episodes for representation learning). We believe \modelshort reduces the barrier to real-world deployment of imitation learning --- especially in more autonomous settings by using the ability to ask for help to mitigate risk.

\paragraph{Limitation/Requirement} In this paper, we rely on the incremental data collection / training scenario, i.e. it needs to be approximately true that each trained policy will be a strict improvement over the previous. This is important because the value approximator is trained using rollouts from previous policies; that is, the learned values are off-policy with respect to the jointly learned Behavioral Cloning policy. As a result, by assuming each new policy is strictly better, the predicted state values learned from previous policies' rollouts form a \textit{lower bound} on the state values for the newer policy.

\printbibliography

@misc{vib,
  doi = {10.48550/ARXIV.2202.07600},
  
  url = {https://arxiv.org/abs/2202.07600},
  
  author = {Du, Yuqing and Ho, Daniel and Alemi, Alexander A. and Jang, Eric and Khansari, Mohi},
  
  title = {Bayesian Imitation Learning for End-to-End Mobile Manipulation},
  
  publisher = {arXiv},
  
  year = {2022},
  
  copyright = {arXiv.org perpetual, non-exclusive license}
}

@misc{momart,
  doi = {10.48550/ARXIV.2112.05251},
  
  url = {https://arxiv.org/abs/2112.05251},
  
  author = {Wong, Josiah and Tung, Albert and Kurenkov, Andrey and Mandlekar, Ajay and Fei-Fei, Li and Savarese, Silvio and Martín-Martín, Roberto},
  
  title = {Error-Aware Imitation Learning from Teleoperation Data for Mobile Manipulation},
  
  publisher = {arXiv},
  
  year = {2021},
  
  copyright = {arXiv.org perpetual, non-exclusive license}
}

@article{Diryag2014NeuralNF,
  title={Neural networks for prediction of robot failures},
  author={Ali Karkara A. Diryag and Marko Mitic and Zoran Miljkovi{\'c}},
  journal={Proceedings of the Institution of Mechanical Engineers, Part C: Journal of Mechanical Engineering Science},
  year={2014},
  volume={228},
  pages={1444 - 1458}
}

@article{AlonsoTovar2016BayesianNC,
  title={Bayesian Network Classifier with Efficient Statistical Time-Series Features for the Classification of Robot Execution Failures},
  author={Jos{\'e} Alonso-Tovar and Cidesi and Baidya Nath Saha and Jesus Romero-Hdz and David Ortega},
  journal={International Journal on Computer Science and Engineering},
  year={2016},
  volume={3},
  pages={80-89}
}

@article{Li2014AnIK,
  title={An Improved Kernel Based Extreme Learning Machine for Robot Execution Failures},
  author={Bin Li and Xuewen Rong and Yibin Li},
  journal={The Scientific World Journal},
  year={2014},
  volume={2014}
}

@article{Liu2019DeepCN,
  title={Deep Convolution Neural Networks for the Classification of Robot Execution Failures},
  author={Yinan Liu and Xiuqing Wang and Xuemei Ren and Feng Lyu},
  journal={2019 CAA Symposium on Fault Detection, Supervision and Safety for Technical Processes (SAFEPROCESS)},
  year={2019},
  pages={535-540}
}

@book{sutton2018reinforcement,
  title={Reinforcement learning: An introduction},
  author={Sutton, Richard S and Barto, Andrew G},
  year={2018},
  publisher={MIT press}
}

@article{Thomas2016DataEfficientOP,
  title={Data-Efficient Off-Policy Policy Evaluation for Reinforcement Learning},
  author={Philip S. Thomas and Emma Brunskill},
  journal={ArXiv},
  year={2016},
  volume={abs/1604.00923}
}

@article{Voloshin2021EmpiricalSO,
  title={Empirical Study of Off-Policy Policy Evaluation for Reinforcement Learning},
  author={Cameron Voloshin and Hoang Minh Le and Nan Jiang and Yisong Yue},
  journal={ArXiv},
  year={2021},
  volume={abs/1911.06854}
}

@article{Fu2021BenchmarksFD,
  title={Benchmarks for Deep Off-Policy Evaluation},
  author={Justin Fu and Mohammad Norouzi and Ofir Nachum and G. Tucker and Ziyun Wang and Alexander Novikov and Mengjiao Yang and Michael R. Zhang and Yutian Chen and Aviral Kumar and Cosmin Paduraru and Sergey Levine and Tom Le Paine},
  journal={ArXiv},
  year={2021},
  volume={abs/2103.16596}
}

@article{Kallus2020DoubleRL,
  title={Double Reinforcement Learning for Efficient Off-Policy Evaluation in Markov Decision Processes},
  author={Nathan Kallus and Masatoshi Uehara},
  journal={J. Mach. Learn. Res.},
  year={2020},
  volume={21},
  pages={167:1-167:63}
}

@inproceedings{Precup2000EligibilityTF,
  title={Eligibility Traces for Off-Policy Policy Evaluation},
  author={Doina Precup and Richard S. Sutton and Satinder Singh},
  booktitle={ICML},
  year={2000}
}

@inproceedings{Precup2001OffPolicyTD,
  title={Off-Policy Temporal Difference Learning with Function Approximation},
  author={Doina Precup and Richard S. Sutton and Sanjoy Dasgupta},
  booktitle={ICML},
  year={2001}
}

@INPROCEEDINGS{ensembledagger,  author={Menda, Kunal and Driggs-Campbell, Katherine and Kochenderfer, Mykel J.},  booktitle={2019 IEEE/RSJ International Conference on Intelligent Robots and Systems (IROS)},   title={EnsembleDAgger: A Bayesian Approach to Safe Imitation Learning},   year={2019},  volume={},  number={},  pages={5041-5048},  doi={10.1109/IROS40897.2019.8968287}}

@inproceedings{Richter2017SafeVN,
  title={Safe Visual Navigation via Deep Learning and Novelty Detection},
  author={Charles Richter and Nicholas Roy},
  booktitle={Robotics: Science and Systems},
  year={2017}
}

@article{Hirose2018GONetAS,
  title={GONet: A Semi-Supervised Deep Learning Approach For Traversability Estimation},
  author={Noriaki Hirose and Amir Sadeghian and Marynel V{\'a}zquez and Patrick Goebel and Silvio Savarese},
  journal={2018 IEEE/RSJ International Conference on Intelligent Robots and Systems (IROS)},
  year={2018},
  pages={3044-3051}
}

@inproceedings{saycan2022arxiv,
    title={Do As I Can and Not As I Say: Grounding Language in Robotic Affordances},
    author={Michael Ahn and Anthony Brohan and Noah Brown and Yevgen Chebotar and Omar Cortes and Byron David and Chelsea Finn and Chuyuan Fu and Keerthana Gopalakrishnan and Karol Hausman and Alex Herzog and Daniel Ho and Jasmine Hsu and Julian Ibarz and Brian Ichter and Alex Irpan and Eric Jang and Rosario Jauregui Ruano and Kyle Jeffrey and Sally Jesmonth and Nikhil Joshi and Ryan Julian and Dmitry Kalashnikov and Yuheng Kuang and Kuang-Huei Lee and Sergey Levine and Yao Lu and Linda Luu and Carolina Parada and Peter Pastor and Jornell Quiambao and Kanishka Rao and Jarek Rettinghouse and Diego Reyes and Pierre Sermanet and Nicolas Sievers and Clayton Tan and Alexander Toshev and Vincent Vanhoucke and Fei Xia and Ted Xiao and Peng Xu and Sichun Xu and Mengyuan Yan and Andy Zeng},
    booktitle={arXiv preprint arXiv:2204.01691},
    year={2022}
}

@misc{tcl,
  doi = {10.48550/ARXIV.2202.01862},
  
  url = {https://arxiv.org/abs/2202.01862},
  
  author = {Khansari, Mohi and Ho, Daniel and Du, Yuqing and Fuentes, Armando and Bennice, Matthew and Sievers, Nicolas and Kirmani, Sean and Bai, Yunfei and Jang, Eric},
  
  title = {Practical Imitation Learning in the Real World via Task Consistency Loss},
  
  publisher = {arXiv},
  
  year = {2022},
  
  copyright = {arXiv.org perpetual, non-exclusive license}
}

@misc{dagger,
  doi = {10.48550/ARXIV.1011.0686},
  
  url = {https://arxiv.org/abs/1011.0686},
  
  author = {Ross, Stephane and Gordon, Geoffrey J. and Bagnell, J. Andrew},
  
  title = {A Reduction of Imitation Learning and Structured Prediction to No-Regret Online Learning},
  
  publisher = {arXiv},
  
  year = {2010},
  
  copyright = {arXiv.org perpetual, non-exclusive license}
}

@ARTICLE{oldMM,  author={Carriker, W.F. and Khosla, P.K. and Krogh, B.H.},  journal={IEEE Transactions on Robotics and Automation},   title={Path planning for mobile manipulators for multiple task execution},   year={1991},  volume={7},  number={3},  pages={403-408},  doi={10.1109/70.88151}}

@INPROCEEDINGS{newMMIL,  author={Welschehold, Tim and Dornhege, Christian and Burgard, Wolfram},  booktitle={2017 IEEE/RSJ International Conference on Intelligent Robots and Systems (IROS)},   title={Learning mobile manipulation actions from human demonstrations},   year={2017},  volume={},  number={},  pages={3196-3201},  doi={10.1109/IROS.2017.8206152}}

@article{newMMIL2,
  title={Combined Task and Action Learning from Human Demonstrations for Mobile Manipulation Applications},
  author={Tim Welschehold and Nichola Abdo and Christian Dornhege and Wolfram Burgard},
  journal={2019 IEEE/RSJ International Conference on Intelligent Robots and Systems (IROS)},
  year={2019},
  pages={4317-4324}
}

@INPROCEEDINGS{dooropeningIL,  author={Nagatani, K. and Yuta, S.},  booktitle={Proceedings of IEEE International Conference on Robotics and Automation},   title={Designing strategy and implementation of mobile manipulator control system for opening door},   year={1996},  volume={3},  number={},  pages={2828-2834 vol.3},  doi={10.1109/ROBOT.1996.506591}}

@article{BCzero,
  doi = {10.48550/ARXIV.2202.02005},
  
  url = {https://arxiv.org/abs/2202.02005},
  
  author = {Jang, Eric and Irpan, Alex and Khansari, Mohi and Kappler, Daniel and Ebert, Frederik and Lynch, Corey and Levine, Sergey and Finn, Chelsea},
  
  title = {BC-Z: Zero-Shot Task Generalization with Robotic Imitation Learning},
  
  publisher = {arXiv},
  
  year = {2022},
  
  copyright = {Creative Commons Attribution 4.0 International}
}

@article{Huber,
author = {Peter J. Huber},
title = {{Robust Estimation of a Location Parameter}},
volume = {35},
journal = {The Annals of Mathematical Statistics},
number = {1},
publisher = {Institute of Mathematical Statistics},
pages = {73 -- 101},
year = {1964},
doi = {10.1214/aoms/1177703732},
URL = {https://doi.org/10.1214/aoms/1177703732}
}

@misc{resnet,
  doi = {10.48550/ARXIV.1512.03385},
  
  url = {https://arxiv.org/abs/1512.03385},
  
  author = {He, Kaiming and Zhang, Xiangyu and Ren, Shaoqing and Sun, Jian},
  
  title = {Deep Residual Learning for Image Recognition},
  
  publisher = {arXiv},
  
  year = {2015},
  
  copyright = {arXiv.org perpetual, non-exclusive license}
}

@article{originalvib,
  doi = {10.48550/ARXIV.1612.00410},
  
  url = {https://arxiv.org/abs/1612.00410},
  
  author = {Alemi, Alexander A. and Fischer, Ian and Dillon, Joshua V. and Murphy, Kevin},
  
  title = {Deep Variational Information Bottleneck},
  
  publisher = {arXiv},
  
  year = {2016},
  
  copyright = {arXiv.org perpetual, non-exclusive license}
}

\end{document}